# Co-Robots as Care Robots

**Oliver Bendel**

School of Business FHNW, Bahnhofstrasse 6, CH-5210 Windisch
oliver.bendel@fhnw.ch

**Alina Gasser; Joel Siebenmann**

F&P Robotics, Rohrstrasse 36, CH-8152 Glattbrugg
ags@fp-robotics.com; jos@fp-robotics.com

**Abstract**

Cooperation and collaboration robots, co-robots or cobots for short, are an integral part of factories. For example, they work closely with the fitters in the automotive sector, and everyone does what they do best. However, the novel robots are not only relevant in production and logistics, but also in the service sector, especially where proximity between them and the users is desired or unavoidable. For decades, individual solutions of a very different kind have been developed in care. Now experts are increasingly relying on co-robots and teaching them the special tasks that are involved in care or therapy. This article presents the advantages, but also the disadvantages of co-robots in care and support, and provides information with regard to human-robot interaction and communication. The article is based on a model that has already been tested in various nursing and retirement homes, namely Lio from F&P Robotics, and uses results from accompanying studies. The authors can show that co-robots are ideal for care and support in many ways. Of course, it is also important to consider a few points in order to guarantee functionality and acceptance.

## Introduction

The cooperation and collaboration robot (short "co-robot" or "cobot") conquers the factories. There it works with its human colleague on a common task (cooperation), possibly hand in hand and in close coordination (collaboration). Whatever it can do faster and better, whatever makes it easier for us, it does it (Bendel 2018c). Production and logistics benefit from lightweight robots, which usually have one arm and five to seven degrees of freedom. Unlike their machine colleagues of conventional design, they do not need a cage or protection. They themselves include the protection, through their programming and their design. ISO and BS standards and other standards state what robots are allowed and not allowed to do; they declare people's eyes and necks to be taboo zones, to name just two examples. However, such co-robots are not completely free in other respects either. They usually do not leave their building and are moved at most from one place to another. Their human counterpart often remains the same for days, or is at least interchangeable. We have a clearly defined, familiar environment, a manageable situation.

In care, support and therapy, one has long been interested in machine helpers, although significant progress has only been made in recent years (Bilyea et al. 2017). Assistance devices appear, from simple tools to intelligent systems (Becker 2018). In therapy, robots exist as products, in care and support mainly as prototypes. There are different types of care robots. They inform, they navigate, they transport something; and they take concrete action, they touch the patients, help them to straighten up, support the nurse or carer to put a patient into another bed (Bendel 2018a). Some prototypes have already disappeared completely; others have become more or less reliable solutions that have changed the context. For example, Riken in Japan has stopped developing Robear for the time being, and the German Fraunhofer IPA robbed of the arms of some Care-O-bots, called them Paul and sent them to shopping centers in Europe, which seems to be worth it (Bendel 2018c).

If body proximity or even body contact is necessary or desired, most devices and robots are not yet perfect. Help with food intake, for example, is a highly complex process in which the machine can easily injure humans, and so are personal hygiene and especially intimate care. Care robots are also rarely independent, and they need the caregivers for a smooth operation. This is not even a disadvantage, because probably nobody wants to be undressed, washed or wrapped in diapers by a soulless machine – only perhaps a profit-driven, unscrupulous business would treat its patients this way.

A solution to some of these challenges could be the above-mentioned co-robots, not as industrial robots, but as service robots (Bendel 2018b). They can cope with the proximity to the patient and meet the high requirements that exist in such an environment (Bendel 2018d). They can be mobilized by being mounted on a platform with wheels or by screwing legs and feet on them. You can get them to take something away, fetch something, hold something or let them open it. They can show how to perform movements; they can massage and stimulate. However, what they can hardly do for the time being is to turn someone around and put him or her in another bed because they are too light and not strong enough. A hand with its two or three fingers can lift two or three kilos. Care, support and therapy are often more complex environments than a factory. Patients come and go just as much as the nurses and carers. If it is a system with natural language skills, special challenges emerge, for example with regard to the design of the voice or the understanding of dialects or accents.

This paper examines how co-robots can be used in care and support and what opportunities and challenges arise. Presentation and discussion refer to Lio, a mobile co-robot that is available in Europe in small series. It was developed by F&P Robotics in 2017 and has been tested in several nursing homes and rehabilitation clinics since that year, partly in collaboration with a university. The specificities of the type and the possible uses are discussed and associated with the results of qualitative studies carried out by the authors. Among others, communication, interaction, movement and design are of interest. This paper works out the potential for co-robots in the role of care robots and what limits arise in practice. Ethical and legal aspects are touched on – but above all, it is about useful hints for human-robot interaction and communication.

## The Service Robot Lio

Lio is a service robot for use in nursing and retirement homes and for people who need support at home. The mobile co-robot, equipped with a soft artificial leather cover, can communicate with people and assist with home requirements and nursing tasks, as you can learn from the website www.fp-robotics.com/de/care-lio/. Technologies of mechatronics and artificial intelligence are used in such a way that Lio is not only helpful, but is also liked and accepted by humans as far as possible.

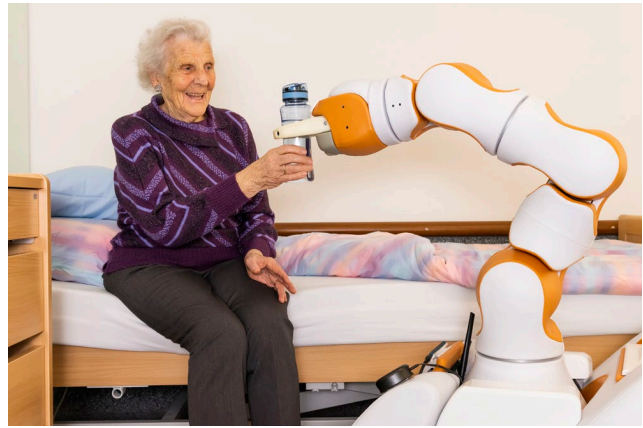

Fig. 1: Lio in action (photo: F&P Robotics)

According to the data sheet, Lio weighs 75 kilograms and is max. 162 centimeters high (F&P Robotics 2019). It has a robotic arm with six degrees of freedom. Its length in outstretched state is about 113 centimeters, that of its gripper 29 centimeters, the maximum payload 3 kilograms. The arm is mounted on a 29 centimeters high mobile platform with four wheels, in which a laser sensor and several ultrasonic sensors are integrated, and can be removed if required. Lio masters autonomous navigation and has path detection (simultaneous positioning and mapping, or SLAM for short) as well as safe avoidance functions. It has storage areas for bottles, cups, plates, etc. A tablet on the platform is used to display the status (emotions, functions, charging, and power display). In addition, a separate operating tablet is available, which can be used by the service worker and through which the robot can be controlled (alternatively, a personal computer or smartphone can be used because it is a web application). Two cameras together with the appropriate software enable person and object recognition as well as gesture recognition. Person recognition is based on face recognition. Speech recognition and output allow the use of natural language. Several loudspeakers and a microphone are installed for the same purpose. A software package includes approximately 30 interaction functions.

Lio can recognize and greet people as well as identify objects, show the current state and possible or upcoming duties, drive around autonomously and charge something, independently grasp something, manipulate and transport as well as receive inputs by voice (English and German), movement or touch. In addition, there are numerous entertainment and interaction functions. The end of Lio's arm is usually equipped with a gripper (see Fig. 1). The standard gripper can be replaced by another end piece, such as variants of grippers (e.g. with vacuum technology) or massage heads. In practical tests, it is repeatedly equipped with two eyes (only magnets reminiscent of eyes), which makes the robot appear animal or human (see Fig. 2).

## Tests and Studies with Lio

Between 2017 and 2019, Lio was tested several times in practice, in nursing and retirement homes as well as in rehabilitation clinics. Various qualitative studies were carried out, within a master thesis at the University of Basel (2017) and in the form of another usability study (2019). Only a small number of people participated in both studies. However, their reactions and statements are interesting and noteworthy, especially since they are directly affected. Large-scale studies in this context are a resource problem and sometimes fail not only because of the withdrawal or prevention of patients, but also because of the reservations of relatives.

In Alina Gasser's master thesis, the use of Lio was evaluated at the Alters- und Pflegeheim Weinfelden (a retirement and nursing home) as well as at the Alterszentrum Bussnang (a retirement centre), both in the canton of Thurgau in Eastern Switzerland. The 18 participants were visited in their private rooms, where the first part of the interview took place. They were then accompanied to the common room of the residents, where, after a short introduction, they solved a few predetermined tasks interacting with the robot, and the details were recorded in a protocol. The second part of the interview then took place. The interviewer, an assistant and the technician were always present. In some cases, caregivers were present as well, watching the interactions from the back of the room. The technician sat behind the participants and steered the robot's driving around. The results were published in a specialist book in 2018 (Früh and Gasser 2018).

The usability study was conducted by F&P Robotics – again under the leadership of Alina Gasser – at two locations, at the Altersheim Agaplesion Bethanien Havelgarten (a retirement home in Berlin, Germany), and at the Rehaklinik Zihlschlacht (a rehabilitation clinic) in the canton of Thurgau, Switzerland. It took place in the personal rooms of the eight test persons and in the meeting rooms or in common areas of the respective institutions. During the study, a technical employee of the company accompanied the robot in order to be able to quickly solve possible dysfunctions during the testing, but above all to start the individual functions of the robot. The results will be published in a specialist book in 2020 (Wirth et al. 2020).

## Dimensions of Use

In the following, the use of co-robots in industry and in care and support is presented based on different dimensions. In each case, the authors discuss the use of Lio, with the help of qualitative studies.

### Safety and Reliability

In the industrial sector, robots are highly regulated. ISO and BS standards apply, among other rules. Some refer specifically to co-robots. One caveat that is often expressed in relation to care robots concerns safety. Depending on the application, the robot comes more or less close to the patients and interacts with them, for example by touching them directly or handing them things or taking these away. Care robots used as prototypes are mostly individual solutions, such as HOBBIT and Robear, and their functions and components must be developed and checked individually in accordance with existing standards (Stahl 2018). In contrast, co-robots have fundamental advantages. Even if they are used in a context other than industry, the principle is tested, proximity to humans is envisaged, and – if they are standardized – functions or components already comply with the specifications.

Lio ensures the safety of the staff and the persons to be cared for by complying with the relevant standards (in particular ISO 13482 Personal Care Robots and ISO 15066 Collaborative Robots) (Früh and Gasser 2018). The approval for Lio in Europe is handled in cooperation with the Swiss National Accident Insurance Fund (SUVA). In addition, the personnel who work with the robot are specifically trained. The provision of security measures, which the nursing staff has to take into account, also plays a role. Lio is able to dodge people and obstacles and has an emergency button that anyone can operate in principle.

There were no safety issues in the above-mentioned practical tests and studies. The patients or caregivers never pressed the emergency button. In addition, complex tasks in which most service robots currently fail, such as "feeding", washing or dressing and undressing, have not been tested. It is, of course, in principle forbidden to expose the probands of a study to a risk. The authors would like to point out that there are fewer obstacles in countries such as China in this respect, and the Swiss-Chinese joint product P-Care is likely to be put into practice under very different conditions.

### Communication

A co-robot in industry mostly has only rudimentary forms of communication. The worker should be able to stop it quickly, which can be solved by tilting an emergency switch, pressing a button on a display or by a voice command. Otherwise, further communication is rarely desired, simply because standardized processes are available and no distraction should take place. In the field of care, this is different. Here, robots have to assert themselves in different situations; they must be able to reach out to users, and users must be able to deal with them. Patients have very different communication skills. Some of them are mentally or physically impaired. The nursing staff sometimes do not speak the national language or do not do so from the beginning. Against this background, it makes sense for a care robot to have several options.

In Lio's case, communication can take place in three different ways, via voice, head and sensors on the grippers (Wirth et al. 2020). Thanks to the first modality, one can answer Lio's questions verbally, and one can ask questions in spoken language. One can press its head (i.e. the end piece of the co-robot) either down to answer Yes, or move it left or right to answer No. Since the arm of a co-robot is very flexible, this form of communication seems to be quite practical. In addition, it can be learned intuitively in our culture, although it could be irritating that it is not one's own head but the head of the counterpart that should be moved. The green and red sensors (each a brightness sensor) can also be used to answer Yes or No. In the following, the authors discuss briefly the natural language communication in practice, but then above all address the touch of the head – i.e. the end piece of the co-robot.

The usability study showed that the test persons often began to communicate orally with Lio on their own initiative as soon as it had spoken, suggesting an intuition of language modality (Wirth et al. 2020). One possible reason could be that the speaking ability is not negatively affected by the physical ailments of many probands. This would imply that a robot in this industry should be able to be controlled by language at any time, thereby minimizing the potentially negative effect of physical limitations on the subjective experience of communication. However, suboptimal speech recognition systems and external noise can significantly affect the reliability of speech recognition, and we therefore recommend a second secondary modality that can be used easily and without a hurdle by each user group.

Lio's secondary communication or interaction modality is related to its head. However, as the usability study showed, test persons in wheelchairs struggled to reach the end piece with their hand (Wirth et al. 2020). For one proband, the distance was too great, which prevented an interaction with Lio while sitting and forced the proband to get up. This physical activity associated with ailments led to discomfort and ultimately to the breakdown of interaction. This implies that the distances between Lio and the various test persons vary too much to ensure a pleasant and easy interaction through a generally accessible predefined position of the head.

The position of a physical modality – this also refers to interactions that are not used for communication – should therefore be able to be adapted to the specific user groups or even individuals so that it is easily accessible to all. An arm movement should suffice. The upper body should normally not have to be moved in the context of care, otherwise, particularly for users with physical discomfort in the upper body, the consequences described above may occur and could ultimately lead to the interruption of the interaction (Wirth et al. 2020). The problem here was not the length of the robot arm, but the fact that it should not protrude from the base surface of the platform. This self-imposed security solution could possibly be softened in a situational manner and with additional security measures.

In a variable position, one should pay attention to possible consistency problems. In Lio's case, according to the usability study, differences in the relative height of the head and neck led to confusion in a test person (Wirth et al. 2020). Usually Lio's highest point is the end piece with the gripper, i.e. its head. In one task, however, it lowered, so that the uppermost link of the neck became the highest point. This was then pressed, as usual, by a test person to confirm a question, which resulted in the input not being registered. Perhaps by chance, by interacting with Lio, the user learned not to press its head to confirm a question, but the highest point of the arm.

**Interaction**

As shown, co-robots work hand in hand with us. They come close to us physically or even touch us (Bendel 2018c). In care, proximity and touch become a must, so to speak. It is about assisting the patient, the elderly or the sick. When the robot is sent away, it is meant to come back with a drug or food. It hands them over to the person concerned, or at least it presents them in such a way that they can grasp them without difficulty. In the future, at least according to the vision of some scientists, companies and media, the robot is supposed to "feed", dress, undress and wash the patient (Becker 2018; Bendel 2018a).

The usability study revealed the need for functions that support test persons in activities they can no longer perform themselves, such as picking up a bottle from the floor and then opening it (Wirth et al. 2020). Additional functions mentioned were the taking down of the food order or the telling of the time. But there was also a demand for entertainment. It can be said that support for activities that the user can no longer perform independently does not seem to be sufficient. A robot should also facilitate repetitive tasks of everyday life and have an entertaining function.

In the future, according to the manufacturers, Lio will be able to take and deliver blood samples, among other things, which will support not only people in need of care, but also nursing staff and medical personnel or auxiliaries. The robot is also intended to gather patients, which is sometimes a task for the nursing staff. Specifically, it should knock on the door (with a special knocking tool), use an ArUco code to find the doorknob, open the door, inform about the appointment, close the door (to maintain privacy) and continue driving.

## Body Movements

One can define body movements as movements that are not forward and backward movements of the entire machine and are not required to complete a task (Früh and Gasser 2018). For example, the bending of the robot and the moving of the grippers around an object are needed to pick it up and are therefore not called body movement. In industry, body movements hardly play a role.

In Lio's case, body movements are used again and again to make it appear more lifelike (Früh and Gasser 2018). Thus, the robot masters a bow when greeting or saying goodbye, can move its torso to look at an object from different angles, can "dance" to the beat of music and swivel its top joint after a question has been asked to attract interest (mimics a tilt of the head, so that the body movements become head positions).

Half of the participants were confronted with such body movements during the evaluation of the master thesis, while the other half were not (Früh and Gasser 2018). Based on the studies of Vincze et al. (2016), Robinson et al. (2014), Tinetti and Williams (1997), Cumming et al. (2000) and Fischinger et al. (2016), the interaction consisted of tasks that were solved together with the robot. Participants taught the robot to recognize an object, indicated whether they wanted to hear a song in return, placed the new object on the ground and ordered the robot to pick it up.

The result of the master thesis was that body movements and head positions contributed to the human appearance and increased user acceptance (Wirth et al. 2020). They also potentially motivated users to tackle the tasks and helped them cope with the tasks.

## Forward and Backward Movements

A co-robot in industry is usually not mobile. Rather, it is firmly attached to a location. There, together with a worker, it takes on predetermined tasks, whereby it should be able to adapt to the different speeds of its human co-worker. It can also learn new actions on site, for example by moving its arm and storing the movement. However, its range is limited to the corresponding radius. This serves not least for occupational safety and assessment.

In care, many applications such as the transport and distribution of drugs and food require the mobility of the co-robot, more concretely forward and backward movements and a rotation around its own axis. For some activities, temporary anchoring in the ground would be useful, for example for local support, re-bedding or straightening up, as this would ensure the necessary stability. For this, however, the robot would have to have a certain robustness and a certain load-bearing capacity that exceeds that of many lightweight robots.

With its four-wheel platform, Lio guarantees mobility. It masters autonomous movement, but it can also be navigated specifically to a desired location using the application on the tablet or smartphone. In practice tests, for example at the Altersheim Agaplesion Bethanien Havelgarten in Berlin, Lio was successfully used for transports and deliveries, namely for snacks (Wirth et al. 2020). Because it can identify individual patients through facial recognition, it can theoretically bring the required drug or food to the right person. However, no errors are allowed, and the functions have to be not only used in further tests, it is necessary to create an environment in practice that excludes errors or delays in delivery. In addition, there are legal restrictions in Germany and Switzerland. In these studies, no evidence has been gained about the forward and backward movements, except that Lio can reliably avoid patients and apologizes to them if it does not have enough space for this.

## Size

A co-robot can vary enormously in height and width, depending on how many degrees of freedom it has and how long its arm is, how often and how strongly it is angled and directed upwards or downwards. For many tasks, the arm is not stretched, but angled several times, and the end piece is relatively far down to assemble or unpack something. However, the object to be manipulated can also be on a pedestal – the worker should reach it comfortably – or lifted by another machine. In the area of care, some patients are in bed, sitting on a chair or in a wheelchair. In these cases, a robot reaches or exceeds them slightly in height.

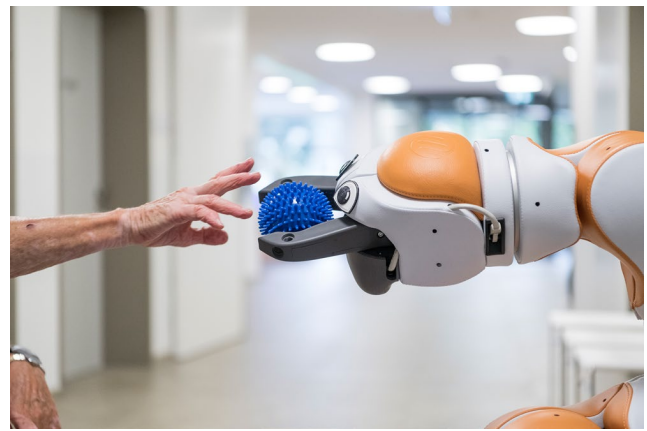

Fig. 2: Lio's eyes (photo: F&P Robotics)

The usability study showed the relative height as a factor affecting the subjectively perceived appearance of a robot. Several test persons described Lio as "terrifying" or "uncanny" if its head was higher or equally high as the proband's head (Wirth et al. 2020). The attitude towards Lio improved, as soon as its highest point was below the eye level of the test persons. This implies that the relative height of the robot compared to the user can have a positive and negative effect on the image of the robot. As a result, the highest point of the robot should be below the user's eye level, so that he or she looks down at the robot and not the other way around and therefore finds it pleasant. In the case of Lio, this finding reinforces the need for an individually adjustable level of secondary modality; it is psychologically, but also ethically relevant.

Another factor is the width of the platform. It should not be too wide, otherwise the robot will not be able to pass through narrow doors or will not have enough space in the elevator next to a wheelchair. However, the platform should not be too small either, because, as with Lio, patients want to hold on to his arm and therefore a certain stability is required. In other words, one must avoid the falling of patients – their well-being is paramount.

### Design

Co-robots do not seem animal- or human-like at first. However, the idea of an animal or human arm arises quickly, and, indeed, it is nothing else than a mechanical arm. If you attach eyes, a nose or a mouth to the end piece, there is a different impression immediately. You think of a snake, a bird or another animal. Depending on the characteristic, a person can also be associated. When you mount two arms, as in ABB's YuMi or F&P Robotics' P-Care, the impression is quite different, and when a body is added and even a real head for the body, a humanoid design in the narrower sense is quickly achieved.

The usability study yielded different results on the question of whether humanoid design is preferred (Wirth et al. 2020). Some test persons found Lio's head with its eyes sweet or cute. However, one of them criticized its eyes, asking why robots must always be human-like. Lio is not human-like (the majority of robots are not), but apparently its eyes make it seem so, although it would have to be examined in more detail whether its overall appearance, including the communication and interaction possibilities, may have led to this impression (Bartneck et al. 2009). In any case, the responsible persons removed the eyes out of consideration for this particular proband. In this context, psychological and ethical questions arise as well.

Obviously, a humanized design can attract both acclaim and displeasure, depending on the user's taste or background. However, the authors can almost rule out a negative influence on the evaluation of Lio's appearance by an Uncanny Valley effect. This should only occur with a very human-like design, a criterion, which Lio obviously does not meet. Robots such as Sophia or Erika, on the other hand, raise high expectations because of their appearance and disappoint with their inhuman, eerie-looking smile (Goddard 2018). Another reason could be an ethical aspect: Some users may find a subjectively perceived human appearance of an assistance robot not justified (Coeckelbergh et al. 2016).

Lio's big eyes may be useful not least in connection with the scheme of childlike characteristics. A female subject in the usability study spoke quite commandingly at the beginning of the interaction, also very clearly and slowly, presumably so that the robot would surely understand her (Wirth et al. 2020). However, that changed over time; the sentences were no longer pronounced over-clearly, the voice became friendlier. Towards the end, the proband treated Lio as if it were a child. At the same time, the negative reactions to Lio's faults as well as its missing but desired functions decreased. A possible explanation for this increase in fault tolerance and user acceptance could be the human qualities of a child that were projected on Lio. This effect could indicate an important design aspect. The robot's failure in a task could be less important due to a scheme of childlike characteristics and would be more likely to be forgiven by the user, which in turn could have a positive effect on the robot's perceived usability.

With a childlike appearance, therefore, the fault tolerance and thus the patience of the user could be increased, which could be particularly advantageous in the observed early termination rate in older people (Giuliani et al. 2005). In addition, a childlike design could possibly reduce the initial mistrust of the user towards the robot (Scopelliti et al. 2005), as it can positively influence the affective reaction to an object (Miesler et al. 2011). Of course, the question is whether a co-robot is a suitable starting point for a childlike design. Robots like Roboy seem childlike without the users having to make an effort to use their imagination (Pfeifer et al. 2013). Finally, a child can also be perceived as inappropriate in the care context, which may weaken acceptance again (Bendel 2018a).

### Personality

Co-robots in industry usually have neither character nor personality. They are not social robots in the stricter sense, even if they work closely with humans. However, this can change if they leave the factory and are used as service robots. At the least, the robot is then perceived more as a counterpart with which one can communicate, for example.

The activities on the manufacturer and supplier side also support this changed perspective. F&P Robotics named their robot Lio. In its early days, for example during the writing of the master thesis, its name was still Angela (Früh and Gasser 2018). With its name, it became a personality and received a gender. The test persons actually reacted to this gender. For example, as part of the usability study, one of them insisted that the robot must have a male voice if it had a male name – clearly, they identified Lio as a male name (Wirth et al. 2020).

Similar to other studies (Broadbent et al. 2009; Syrdal et al. 2008; Walters et al. 2008) the participants expected in the survey of the master thesis that the robot should have a certain personality (Früh and Gasser 2018). They wanted to be surprised with the information, which the robot's favorite song is. They wanted to help him learn and develop. The robot should be intelligent, social and spontaneous. Qualitative observations showed that the participants imitated Lio's gestures. They bowed when it bowed, and when it danced to its favorite music, the participants sometimes imitated its lateral movements in the chair or swayed their feet with the beat. These results are certainly worth another study with a larger sample.

**Surveillance**

In principle, one can use a co-robot in the industry to supervise the employee (Bendel 2018c). In fact, however, this is unlikely to be the case. Of course, the employee is forced into a certain routine, and as with the use of all digital devices, the type of use can be traced in principle. However, there are no known cases where the cameras and sensors were used for surveillance in the proper sense. The fixed cameras on the walls and ceilings of offices and factories – quite common in the US – present a greater threat to informational autonomy.

The co-robot in care and therapy becomes more mobile and flexible as it meets different situations and people. This means that in a sense it must be more mature than other models. Cameras, radar, lidar and ultrasonic systems, sensors and actuators of all kinds are required for trouble- and accident-free forward and reverse movement and personal all-round support. Care-O-bot and Pepper have facial and speech recognition, by the way. The emotional robot from Aldebaran or SoftBank, which supports in the home, in shopping malls as well as in nursing and retirement homes, can even identify the individual voice and react to it. Co-robots can be equipped with voice recognition in the same way, and some of them have facial and speech recognition when used as service robots. This, in turn, means that the person to be cared for can be monitored. Visual and auditory data are merged in a never-ending stream that is evaluated by systems and potentially used by humans. In principle, they are even able to observe and listen to adolescent or adult patients directly via cameras and microphones (Bendel 2014).

The area of care and therapy is a private or semi-public environment in which co-robots can compromise privacy. In addition, informational autonomy is affected. This ethical concept has its equivalent on the legal side with "informational self-determination". One should be able to have sovereignty over one's personal data, to be allowed data insight, to prevent the dissemination of data, to be able to force the deletion of data. All this becomes difficult in a room that is shared with robots and where, for safety reasons alone, hardly a door is allowed to remain closed (Früh and Gasser 2018). The described gathering of patients, in which closing the door is an important action, is not self-evident and technically demanding.

Lio is quite capable of gaining sensitive data via facial and speech recognition. The camera can basically take pictures of the patients and can theoretically be abused. The authors discussed the use of Lio with the institutions with regard to data protection and informational autonomy. Patients were informed about the possibilities of the robot and in particular its camera. The people responsible handled the data with care. However, abuse cannot be ruled out in the future. It is in the interest of the company to take into account as many stakeholders as possible, and in the interest of all stakeholders to reach a consensus. The starting point can be a patient decree as proposed by Bendel (2018a) which can prohibit or restrict the use of a care robot.

**From One Arm to Two Arms**

In 2014, ABB launched a two-arm co-robot called YuMi targeting the electronics industry. This design has some advantages and a few drawbacks. The robot can hold a thing in one hand and manipulate it with the help of the other. It can enfold something, from both sides, and lift a large or long object. The arms create a certain impression on the spectator, have a human or animal-like effect on them, raise different expectations and fears depending on the culture. You think of hugging, lifting up – or a headlock. In principle, the robot becomes heavier and more room-filling. A one-armed robot is an elegant solution. Nevertheless, in some contexts, the two-armed model probably belongs to the future.

At WRC 2018 in Beijing, F&P Robotics presented a two-armed co-robot that can be used in therapy and care. P-Care is mobile and reminiscent of a monkey; it is based on the Monkey King (Früh and Gasser 2018). A two-armed co-robot could lead to applications in this area that were not previously in focus or that overwhelmed the prototypes. Care-O-bot and Robear also have two extremities and many skills in dealing with us. However, they hardly have the years of experience gained with co-robots in the industry. As a promotional video shows, Care-O-bot can present a rose, but whether it can administer food to a patient without injuring them remains to be seen. Exactly this could be, as mentioned above, the strength of co-robots such as the product by F&P Robotics: they could "feed" a patient by fixing the head with one hand and with the other bring the spoon to the mouth; they could massage like a human and even help a man urinate. Moreover, they could fetch bulky objects without damaging them. So far, there are no studies in Europe on P-Care. In principle, one must examine the two-armed robot from a psychological and ethical perspective.

## Summary and Outlook

This article presented the advantages, but also the disadvantages of co-robots in care and support. It based its analysis on a model that has already been tested in various facilities, Lio by F&P Robotics, and the results of two qualitative studies that have been conducted are cited. It turns out that co-robots like Lio are ideal for the demands of care and support in many ways. Of course, there are also a few points worth considering to ensure functionality and acceptance.

Time and again, there were challenges that no one had thought of beforehand, for example with regard to operation or effect. This suggests that one should consult further studies with comparable scenarios more thoroughly, and, on the other hand, one should further test the practical use of co-robots in care and accompany them with scientific methods. The cited studies were able to provide initial evidence. However, they have to be conducted with more test persons, and qualitative findings must be followed by quantitative findings. In addition, the findings have to be evaluated not only from human-robot interaction and communication, but also from the point of view of psychology and ethics – and must in turn be incorporated into the technical design.